\documentclass[mnsc]{informs3}
\DoubleSpacedXI

\usepackage{amssymb,amsmath,amsxtra,amstext,xfrac,mathtools}
\usepackage{graphicx,ccaption,booktabs,enumerate,paralist,mdwlist}
\usepackage{dsfont,verbatim,latexsym}
\usepackage{bbm}
\usepackage{bm}
\usepackage{longtable,multirow,threeparttable}
\usepackage{float}
\usepackage{makecell}
\usepackage{eurosym}
\usepackage[english]{babel}
\usepackage{fancyhdr}
\usepackage{xcolor}
\usepackage{booktabs}
\usepackage{caption}
\usepackage{subcaption}

\usepackage{algorithm,algorithmic}
\usepackage{pifont}

\usepackage[T1]{fontenc} 
\usepackage{lmodern} 
\usepackage{fix-cm}

\usepackage[colorlinks=TRUE,citecolor=MyDarkBlue,urlcolor=MyDarkBlue,linkcolor=MyDarkBlue]{hyperref}
\definecolor{MyDarkBlue}{RGB}{158,0,0}

\usepackage{soul}
\sethlcolor{lightgray}

\usepackage[most]{tcolorbox} 
\lstdefinestyle{graybox}{
  backgroundcolor=\color{gray!10},
  basicstyle=\ttfamily\small,
  frame=single,
  breaklines=true,
  columns=fullflexible,
  keepspaces=true,
  showstringspaces=false,
  numbers=left,
  numberstyle=\tiny\color{gray},
  numbersep=10pt
}

\makeatletter
\newcounter{promptbox}
\renewcommand{\thepromptbox}{\arabic{promptbox}}

\newtcolorbox{promptbox}[1][]{
  enhanced,
  breakable,
  colback=white,
  colframe=black,
  colbacktitle=white,
  coltitle=black,
  boxrule=0.5pt,
  arc=1mm,
  left=5pt,
  right=5pt,
  top=5pt,
  bottom=5pt,
  fonttitle=\bfseries,
  fontupper=\ttfamily\sloppy, 
  before skip=10pt,
  after skip=10pt,
  toptitle=3pt,
  bottomtitle=3pt,
  before title={\refstepcounter{promptbox}\phantomsection},
  before upper={\edef\@currentlabel{\thepromptbox}},
  #1
}
\makeatother


\usepackage[sort]{natbib}
 \bibpunct[, ]{(}{)}{,}{a}{}{,}%
 \def\bibfont{\footnotesize}%

\TheoremsNumberedThrough    
\ECRepeatTheorems
\EquationsNumberedThrough

\usepackage{booktabs}

\newcommand{\cmark}{\ding{51}} 
\newcommand{\xmark}{\ding{55}} 

\renewcommand{\theARTICLETOP}{}

\begin{document}



\RUNTITLE{Improving Behavioral Alignment}
\RUNAUTHOR{Kong et al.}

\TITLE{
Improving Behavioral Alignment in LLM Social Simulations via Context Formation and Navigation
}
\ARTICLEAUTHORS{%
\AUTHOR{Letian Kong}
\AFF{CUHK Business School, The Chinese University of Hong Kong, \EMAIL{kongletian@link.cuhk.edu.hk}}

\AUTHOR{Qianran(Jenny) Jin}
\AFF{CUHK Business School, The Chinese University of Hong Kong, \EMAIL{ jennyjin@cuhk.edu.hk}}

\AUTHOR{Renyu(Philip) Zhang}
\AFF{CUHK Business School, The Chinese University of Hong Kong, \EMAIL{ philipzhang@cuhk.edu.hk}}

} 

\ABSTRACT{%

Large language models (LLMs) are increasingly used to simulate human behavior in experimental settings, but they systematically diverge from human decisions in \textit{complex decision-making environments} where participants must anticipate others' actions and form beliefs based on observed behavior. We propose a two-stage framework for improving behavioral alignment. The first stage, \textit{context formation}, explicitly specifies the experimental design to establish an accurate representation of the decision task and its context. The second stage, \textit{context navigation}, guides the reasoning process within that representation to make decisions. We validate this framework through a focal replication of a sequential purchasing game with quality signaling \citep{kremer2016inferring}, extending to a crowdfunding game with costly signaling \citep{cason2024signaling} and a demand-estimation task \citep{gui2025challenge} to test generalizability across decision environments. Across four state-of-the-art (SOTA) models (GPT-4o, GPT-5, Claude-4.0-Sonnet-Thinking, DeepSeek-R1), we find that complex decision-making environments require both stages to achieve behavioral alignment with human benchmarks, whereas the simpler demand-estimation task requires only \textit{context formation}. Our findings clarify when each stage is necessary and provide a systematic approach for designing and diagnosing LLM social simulations as complements to human subjects in behavioral research.

}%





\maketitle

\section{Introduction}\label{sec:intro}

Generative artificial intelligence (AI) and large language models (LLMs) are opening new possibilities for studying human behavior in social science and business research \citep{kane2025augmented, wiberg2025synergizing, gopal2025inventing, demszky2023using, bail2024can}. A particularly promising application is using LLMs to simulate human decision-making in experiments, an approach known as \textit{LLM social simulations}. In these simulations, LLMs act as synthetic subjects, making it possible to study behavioral patterns and evaluate interventions at unprecedented scale \citep{anthis2025llm, charness2025next, manning2025general, toubia2025database}. These simulations offer clear advantages as they are fast, reproducible, and inexpensive compared to recruiting human participants. More importantly, they enable research in settings where human experiments are costly, impractical, or impossible. Recent research explores applications in market research and related evaluation tasks, such as preference measurement and perceptual analysis \citep{brand2024using, li2024frontiers, wang2024large}, while industry pilots are already emerging in automated experimentation \citep{wang2025agenta} and synthetic customer research.\footnote{Examples include Delve AI (\url{https://www.delve.ai/}), which generates synthetic customer personas for market research.} As LLM social simulations expand from research settings into practice, reliable \textit{behavioral alignment}, defined as the extent to which LLM-simulated subjects' decisions correspond to those of human participants, becomes increasingly critical for both scientific validity and practical deployment.

Despite this importance, fundamental questions about when LLM social simulations reliably match human behavior remain unanswered. Early evidence from psychology and economics suggests that LLMs closely reproduce aggregate human response patterns in relatively simple settings, such as trust games and personality assessments \citep{aher2023using, chen2023emergence, toubia2025database}. However, this apparent success breaks down in \textit{complex decision-making environments} characterized by strategic interdependence and endogenous belief formation, where participants must anticipate others' actions and form beliefs based on observed behavior. In these environments, LLMs often diverge from human decisions \citep{akata2025playing, gao2024take}, and LLM-based replications of established experimental findings yield inconsistent results \citep{kirshner2024artificial, chen2025predicting}. Notably, these alignment failures arise precisely in the complex decision-making environments where human experiments are most difficult and costly to conduct, thereby weakening the central motivation for using LLM social simulations in the first place.

Two broad approaches have been used to improve the alignment of LLM social simulations, but neither provides a systematic solution for complex decision-making environments. First, data-driven approaches embed human data directly into LLM social simulations through fine-tuning or calibration. Simulations show substantial gains when models are calibrated to observed human choices \citep{manning2025general} or trained on large-scale behavioral data \citep{binz2025foundation}. However, the effectiveness of fine-tuning depends critically on the match between training tasks and target applications, as models can fail to generalize across different decision environments \citep{gao2024take, toubia2025database}. More fundamentally, reliance on human data creates a bottleneck in the complex decision-making environments where simulations are most valuable, as such data are costly, task-specific, and often unavailable. 

Second, prompt-based approaches use prompt engineering to improve simulation alignment without additional human data. Widely used strategies include Chain-of-Thought (CoT) reasoning \citep{kojima2022large, wei2022chain} and persona adjustments \citep{aher2023using, xie2024can, kirshner2024artificial, li2024frontiers}. While these approaches can improve performance in individual experimental replications, accumulated evidence shows that their effects are inconsistent across tasks and models \citep{goli2024frontiers, leng2023llm}. Some studies go beyond generic prompting strategies and propose targeted interventions for particular decision environments. For example, alignment can be improved by prompting models to explicitly reason about others' likely actions in repeated games \citep{akata2025playing}, or by making the experimental design explicit to the model in order to reduce confounding in LLM-based simulations \citep{gui2025challenge}. However, these task-specific solutions do not generalize across different decision environments, and we still lack a systematic framework for understanding when different approaches succeed or fail.

The main goal of this paper is to examine the sources of behavioral misalignment when LLMs simulate human decision-making in complex decision-making environments. Motivated by the widely observed divergence between LLM-simulated subjects and human behavior in such settings, despite the growing capabilities of state-of-the-art (SOTA) LLMs, we seek to address the following key research question: \textit{How can behavioral alignment in LLM social simulations of complex decision-making be systematically diagnosed and improved}?

To answer our research question, we develop a two-stage theoretical framework inspired by human problem-solving \citep{simon1971human, newell1972human}. The first stage, \textit{context formation}, explicitly specifies the experimental design to establish an accurate representation of the decision task and its context. The second stage, \textit{context navigation}, guides the reasoning process within that representation to make decisions. We empirically evaluate this framework through simulations of three established experimental studies that vary in strategic interdependence and endogenous belief formation. By systematically varying whether prompts provide \textit{context formation}, \textit{context navigation}, or both, we isolate the contribution of each stage to behavioral alignment. We examine models' CoT reasoning traces to diagnose the sources of misalignment. Comparing LLM-simulated subjects' behavior to human benchmarks allows us to assess when and in what ways each stage contributes to behavioral alignment across decision environments and models. We highlight our contributions as follows.

\subsection{Theoretical Framework for LLM Decision-Making}\label{sec:intro_framework}
As our main contribution, we propose a theoretical framework that conceptualizes behavioral alignment between LLMs and humans as a two-stage process involving \textit{context formation} and \textit{context navigation}. By explicitly separating these stages, our framework provides a holistic and theoretically motivated perspective for understanding and improving behavioral alignment across decision environments.

This separation is critical because it reveals a hierarchical structure. \textit{Context navigation} is effective only when grounded in accurate \textit{context formation}. In complex decision-making environments, providing LLMs with task instructions alone leaves both task representation and the reasoning strategy ambiguous. As a result, models must infer both from patterns learned during pretraining, and the resulting representations and reasoning strategies can systematically differ from human approaches. This hierarchical relationship means that misalignment can stem from inadequate \textit{context formation}, from inappropriate \textit{context navigation} within an otherwise accurate representation, or from failures in both stages. Importantly, this hierarchy explains why generic reasoning prompts such as CoT often show inconsistent effects because they attempt to improve navigation while explicitly assuming a task representation that may not exist. Our framework systematically addresses both stages, with particular emphasis on establishing accurate \textit{context formation} as the necessary foundation for effective behavior alignment.

\subsection{Empirical Validation of Our Framework}\label{sec:validation}
To demonstrate the practical scope of our framework, we empirically validate its generalizability along two dimensions: decision environments and model choices. Across decision environments, we apply the framework to three established experimental studies, including two complex behavioral experiments characterized by strategic interdependence and endogenous belief formation, and a simple demand estimation study driven by the law of demand. Across models, we implement a common simulation procedure and evaluation protocol using multiple SOTA LLMs from different model families, including GPT-4o, GPT-5, Claude-4.0-Sonnet-Thinking, and DeepSeek-R1.

Across these settings, our results reveal a consistent pattern: in complex decision-making environments, achieving behavioral alignment requires both \textit{context formation} and \textit{context navigation}, whereas in simple decision environments, \textit{context formation} alone is sufficient. These patterns are robust across models, demonstrating that the framework generalizes across both decision environments and model choices. In addition, we examine models' CoT reasoning traces to diagnose the sources of misalignment, moving beyond outcome comparisons to understand how models interpret tasks and reach decisions.

Our study provides two key insights for designing reliable LLM social simulations. First, the need for explicit reasoning guidance depends on decision environments. In simple, non-strategic settings, making the task representation explicit and accurate is often sufficient for alignment, whereas strategically interdependent and belief-driven environments require additional guidance on how to reason within the representation. Second, while advances in LLM reasoning capabilities \citep[e.g.,][]{guo2025deepseek} can reduce reliance on explicit \textit{context navigation} guidance, they do not eliminate the need for careful \textit{context formation}. Together, our findings highlight that misalignment in LLM social simulations can stem from coherent reasoning applied to inaccurate task representations, with the relative importance of each stage depending on task complexity.

The rest of this paper is organized as follows. In Section~\ref{sec:literature}, we review the relevant literature. Section~\ref{sec:framework} develops our theoretical framework based on human problem-solving theory. In Section~\ref{sec:simulation_empirical}, we present our focal empirical application \citep[][]{kremer2016inferring} and evaluate the alignment with human benchmarks. Section~\ref{sec:applications} applies our framework to two additional experiments \citep[][]{gui2025challenge,cason2024signaling} and demonstrate its generalizability across multiple SOTA models.  Section~\ref{sec:conclusion} concludes.

\section{Literature Review}\label{sec:literature}

Our study is closely related to three streams of research: (i) generative AI and LLM in business applications, (ii) the use of LLMs to simulate human subjects in experiments and simulations, and (iii) alignment challenges and solutions in LLM social simulation.


\subsection{Generative AI and LLM in Business Applications}
\label{sec:lit1}

Generative AI has introduced a new class of technologies capable of producing text, code, designs, and other content at scale, spurring a rapidly growing literature on its economic and social implications. In work settings, access to LLMs and their outputs can accelerate task completion and, in many contexts, improve output quality and productivity at both the individual and organizational levels \citep{noy2023experimental, dell2023navigating, peng2023impact, brynjolfsson2025generative}. For creative tasks, LLM-generated ideas and content tend to increase individual creative output, but may reduce diversity and novelty in collective outputs \citep{doshi2024generative, hou2025double}. On online platforms, AI-generated content reshapes voluntary contribution and participation dynamics, often concentrating activity among higher-capability contributors \citep{burtch2024consequences,shan2025examining}, whereas AI-generated metadata (e.g., titles and hashtags) significantly boosts user engagement by improving user-content matching and inspiring better human-AI co-creation \citep[][]{zhang2024value}. At the labor-market level, the diffusion of generative AI is associated with observable changes in labor demand and employment outcomes, with evidence that the magnitude and direction of these effects vary across occupations and depend on how AI is adopted in work settings \citep{eloundou2023gpts, hui2024short, demirci2025ai, hu2025balancing}.

Beyond studying the impacts of generative AI, a parallel research stream develops LLM-based methods as tools to empower business applications and research. Generative models enhance content production workflows by increasing the productivity and scalability of text and visual content creation in marketing and communication, while also introducing the risk of greater homogenization of marketing content across firms \citep{reisenbichler2022frontiers, reisenbichler2025applying, hackenburg2024evaluating,liu2025generative}. They are also embedded in decision support and process automation, where generated evaluations and classifications feed into systematic processes for strategic analysis, experimentation, and professional knowledge summarization \citep{guo2024automated, liu2025automating, ye2025lola, doshi2025generative}. Related work examines how work is split between humans and LLMs, showing that different ways of using models alongside humans lead to different creative, innovation, and research outcomes \citep{boussioux2024crowdless, chen2024large, arora2025ai, cillo2025generative}. Other studies integrate LLMs into research workflows to support conceptual development, systematic analysis, and scalable data construction in behavioral research \citep{grossmann2023ai, kim2023ai, zhou2024can, carlson2024use, cheng2024human}.

Within this broad research landscape on generative AI in business, LLMs have demonstrated substantial generative and reasoning capabilities across diverse application domains. An open question is how these capabilities translate when models are used to produce judgments and decisions to resemble those of humans.

\subsection{LLM Social Simulations}
\label{sec:lit2}

As LLMs become more prevalent in business and academia, researchers are examining whether they can complement or even substitute for human participants in surveys and lab experiments. The idea is to use LLMs to generate judgments, behaviors, and choices that would otherwise require recruiting human subjects, offering a scalable and low-cost alternative to traditional social science and business research \citep{grossmann2023ai, anthis2025llm}. Early evidence from \citet{argyle2023out}, who introduce the notion of ``silicon samples'', shows that LLMs can closely reproduce the response distributions observed in human surveys.

A growing body of research has since examined the extent to which these silicon samples resemble human subjects in their behavioral and psychological responses. In relatively simple experimental settings, LLMs have been shown to reproduce fundamental human traits such as fairness and reciprocity in classic games \citep{aher2023using, xie2024can, mei2024turing}, align closely with human moral consensus \citep{dillion2023can}, emulate behaviors such as altruism and risk aversion \citep{filippas2024large}, and display human-like cognitive biases \citep{salecha2024large, chen2025manager}. Consistent with these findings, large-scale replication efforts report high success rates, including replication of approximately 75\% of 156 psychological and management experiments \citep{cui2024can} and eight out of nine classic behavioral-operations experiments \citep{kirshner2024artificial}. Beyond replication, several studies propose methodological frameworks that integrate LLM-simulated subjects into research workflows to support theory development and hypothesis testing \citep{manning2024automated, tranchero2024theorizing, ludwig2025large}, and deploy them in broader society and system-level simulations \citep{yang2024oasis, park2022social, chuang2024simulating, yang2025twinmarket, allouah2025your}. 

Taken together, this stream of research suggests that LLMs can reproduce many well-established human behaviors across tasks with clearly specified decision problems and limited ambiguity in how information maps to choices. However, human decision-making often unfolds in multi-agent environments characterized by strategic interdependence and endogenous belief formation, which we refer to as complex decision-making environments. It remains unclear how well LLMs capture human behaviors in such settings. Our study contributes to this literature by systematically evaluating LLM-simulated subjects in these environments.

\subsection{LLM Alignment Challenges and Solutions}
\label{sec:lit3}

To effectively leverage LLMs in research and practical applications, alignment, the extent to which model outputs correspond to human intentions, preferences, and values, has been a central challenge. In LLM social simulations, this concern becomes whether alignment mechanisms developed in computer science are sufficient to ensure the generation of human-like behavior in complex decision-making environments. Existing approaches primarily operate either at training or inference stages. Training-time methods, including supervised fine-tuning \citep[SFT,][]{wei2021finetuned, ouyang2022training}, reinforcement learning with human feedback \citep[RLHF,][]{christiano2017deep, ziegler2019fine}, and reinforcement learning with verifiable rewards \citep[RLVR,][]{lambert2024tulu, guo2025deepseek}, shape model behavior by encouraging adherence to natural-language instructions, human preferences, and verifiable reasoning objectives. At the inference time, methods relying on careful and systematic prompt design such as Chain-of-Thought \citep[CoT,][]{wei2022chain, kojima2022large}, synergizing reasoning and action \citep[ReAct,][]{yao2022react}, and tree-of-thoughts \citep[ToT,][]{yao2023tree} guide models towards more reliable reasoning across diverse tasks. 

In LLM social simulations, behavioral alignment, the extent to which LLM-simulated subjects' decisions correspond to those of human participants, remains a central bottleneck despite advances in training-time and inference-time alignment methods. Relative to human benchmarks, LLMs often exhibit higher levels of rationality \citep{chen2023emergence, kirshner2024artificial}, and less stable intertemporal preference structures \citep{goli2024frontiers}. Misalignment is even more pronounced in complex decision-making environments where participants need to anticipate others' actions and form beliefs based on observed behavior. In such settings, LLMs tend to form systematically biased beliefs and exhibit cooperation patterns and reasoning processes that diverge from those of humans \citep{brookins2023playing, bauer2025can, gao2024take}. While some studies go beyond documenting misalignment by providing diagnostic analyses of its sources, most stop short of proposing generalizable strategies for improving alignment.

Closest to our paper are a few recent studies that propose targeted interventions to improve behavioral alignment in specific settings. For example, \citet{akata2025playing} show that coordination behavior can be improved through a social CoT prompting strategy that encourages models to anticipate others' actions. \citet{gui2025challenge} demonstrate that standard blinded experiment designs can induce confoundedness in LLM-based simulations and propose unblinding as a remedy. Other approaches incorporate human data more directly. \citet{binz2025foundation} introduce \textit{Centaur}, an LLM fine-tuned on large-scale human behavioral data from psychological experiments. \citet{manning2025general} combine theory-grounded mechanisms with human data from related experiments to predict behaviors in novel strategic environments. While effective within their respective domains, these approaches remain task-specific and do not offer a unified theoretical account of why LLMs diverge from human behavior in complex decision-making.

Building on this literature, we adopt a decision-theoretic perspective that shifts attention from task-specific fixes to the underlying decision processes that lead to behavioral alignment. Rather than focusing on procedural and/or technical adjustments alone, we examine how LLMs form representations of decision tasks and reason with those representations to make decisions. We propose a framework that diagnoses the sources of behavioral misalignment in complex decision-making environments and provides a systematic approach for addressing it. By doing so, our approach connects principles from human cognition and social decision-making to the methodological design of LLM social simulations.

\section{Theoretical Framework}\label{sec:framework}


To develop this framework for LLM-based simulations of human decision-making behaviors, we draw on \citet{simon1971human}'s theory of human problem-solving. 

\subsection{Human Problem-Solving}\label{sec:human_problem_solving}
 
Research on human problem-solving shows that people usually make decisions in two stages \citep{simon1971human, newell1972human}. In the first stage, they form an internal representation of the decision task by clarifying what the question is asking, what information is relevant, and what the goal of the problem is. This task representation provides the structure within which the task can be understood. In the second stage, people reason within this representation to determine how to solve the problem, evaluating potential actions using heuristics that guide their search. Empirical studies further support this two-stage view of problem-solving. Changes in how a problem is described lead people to form different task representations  \citep{hayes1974understanding}. Furthermore, because people navigate problems through selective search rather than exhaustive evaluation \citep{newell1972human, bhaskar1977problem}, changes in task representation shape the reasoning process and can thus produce different solution paths even when the underlying task remains the same \citep{simon1976understanding}.

Formally, we define the query $Q$ as the externally provided description of the task (e.g., the instructions and information shown to participants), the context $C_{\text{H}}$ as the internal representation of the decision task, and the navigation strategy $N_{\text{H}}$ as the selective reasoning procedure used to navigate within this representation as . For example, consider a sequential purchasing task, the query $Q$ corresponds to the written instructions shown to a participant in a given decision round, such as describing the product, the observed wait time, and the payoff structure. The context $C_{\text{H}}$ captures how the participant interprets this information. For example, whether they understand wait time as a potential signal of product quality or as merely a cost. The navigation strategy $N_{\text{H}}$ reflects how the participant reasons within this context, such as whether they update beliefs about quality based on prior purchases or rely on a simple expected-value comparison.

While both contexts and navigation strategies can in general be viewed as draws from underlying distributions reflecting heterogeneity in interpretation and reasoning, in controlled experimental settings, these distributions are typically tightly concentrated around an intended task representation and solution pathway, which we denote by $C_{\text{H}}$ and $N_{\text{H}}$. Given $(Q,C_{\text{H}},N_{\text{H}})$, the human response is generated by sampling from a distribution:
\[
Y_{\text{H}} \sim f_{\text{H}}(Q, C_{\text{H}}, N_{\text{H}}),
\]
where $f_{\text{H}}$ is a random function that captures how the query, the context, and the navigation strategy jointly generate the observed human decision $Y_{\text{H}}$.

In practice, researchers use procedures such as piloting and manipulation checks to ensure that participants form the intended representation of the task rather than idiosyncratic ones. These procedures validate the context $C_{\text{H}}$ constructed from $Q$ and concentrate its distribution around the intended interpretation. Once this validated context is in place, the navigation strategy $N_{\text{H}}$ tends to follow systematically from the task representation, allowing human responses to reflect a largely shared understanding of the query $Q$.

Although human cognition and LLM prediction rely on different underlying mechanisms, 
both involve forming task representations and reasoning within those 
representations. Accordingly, the theory of human problem-solving provides a 
foundation for understanding when and why LLM social simulations misalign with human 
behavior, and for designing interventions to improve alignment.

\subsection{Context Formation and Context Navigation}\label{sec:context_formation_navigation}

Motivated by the human problem-solving process \citep{simon1971human, newell1972human}, we propose a two-stage framework designed to diagnose the root causes of misalignment in LLM simulations of human behaviors and to effectively improve human-LLM behavioral alignment. This framework identifies and addresses two key challenges that could cause misalignment: \textit{context formation} and \textit{context navigation}. \textit{Context formation} refers to how LLMs form an accurate task representation, whereas \textit{context navigation} refers to how LLMs reason within that context to generate decisions consistent with human behaviors. Once we specify the query $Q$, the context $C$, and the navigation $N$, an LLM will generate the response as
\[
Y \sim f_{\text{LLM}}(Q, C, N),
\]
where $f_{\text{LLM}}$ is a random function powered by LLM that generates output based on the input prompt of the triplet $(Q,C,N)$.

SOTA LLMs such as GPT-5\footnote{OpenAI. (2025). Introducing GPT-5. Retrieved from \url{https://openai.com/index/introducing-gpt-5/}.} and DeepSeek-R1 are aligned with human behaviors in a wide range of tasks such as math, coding, and commonsense Q\&As \citep[e.g.,][]{guo2025deepseek}.  It is therefore reasonable to assume that $f_{\text{LLM}}$ well approximates $f_{\text{H}}$ with the same input $(Q,C,N)$. In particular, if the LLM has the same input as the human, $Y \sim f_{\text{LLM}}(Q, C_{\text{H}}, N_{\text{H}})$ will follow a similar distribution to the human response $Y_\text{H}\sim f_{\text{H}}(Q, C_{\text{H}}, N_{\text{H}})$. In this case, $f_{\text{LLM}}$ is a simulator that could credibly replicate human behaviors regarding query $Q$. This assumption has also been validated by various successful LLM-based social simulations \citep[e.g.,][]{kirshner2024artificial,cui2024can,mei2024turing}.

However, if only the query $Q$ is provided to the LLM but the context $C$ or the navigation $N$ is missing, the model will automatically infer such information from its prior knowledge. We refer to this setting as the \textit{baseline} case. More specifically, the LLM, acting as a simulator, will first sample the context $C_{\text{Base}}$ based on the query $Q$, then sample the navigation $N_{\text{Base}}$ based on the query and the context $(Q,C_{\text{Base}})$, and finally generate the outcome $Y_{\text{Base}}$ based on the query and the sampled context and navigation, i.e.,
\begin{equation}\label{eq:base-simulator}
C_{\text{Base}}\sim \mathbb{P}_{\text{LLM}}(C|Q),\mbox{ },N_{\text{Base}} \sim \mathbb{P}_{\text{LLM}}(N|Q,C_{\text{Base}}),\mbox{ }Y_{\text{Base}}\sim f_{\text{LLM}}(Q,C_{\text{Base}},N_{\text{Base}}),
\end{equation}
where $\mathbb{P}_{\text{LLM}}(C|Q)$ is the distribution of $C$ conditioned on $Q$ and $\mathbb{P}_{\text{LLM}}(N|Q,C)$ is the distribution of $N$ conditioned on $(Q,C)$, both of which capture the prior knowledge of the pretrained model rather than the specific problem setting. In general, the sampled context and navigation $(C_{\text{Base}},N_{\text{Base}})$ deviate from those in the real human setting $(C_{\text{H}},N_{\text{H}})$. As a consequence, $Y_{\text{Base}}$ is unlikely to be aligned with $Y_{\text{H}}$.

The setting in which the query $Q$ and the accurate context humans perceive $C_{\text{H}}$ are specified, while the navigation strategy remains unspecified, is referred to as the \textit{context formation} case. In this case, the model will infer the navigation strategy, $N_{\text{CF}}$ based on $(Q,C_{\text{H}})$, and generate the outcome $Y_{\text{CF}}$ based on $(Q, C_{\text{H}}, N_{\text{CF}})$, i.e.,
\begin{equation}\label{eq:formation-simulator}
N_{\text{CF}} \sim \mathbb{P}_{\text{LLM}}(N \mid Q, C_{\text{H}}),\mbox{ }Y_{\text{CF}}\sim f_{\text{LLM}}(Q, C_{\text{H}}, N_{\text{CF}}).    
\end{equation}
Compared with the \textit{baseline} case, context formation improves alignment via specifying the accurate context $C_{\text{H}}$. Yet, misalignment may still occur if the LLM-generated navigation strategy $N_{\text{CF}}$ deviates from the human navigation $N_{\text{H}}$, thus yielding the simulated outcome $Y_{\text{CF}}$ misaligned with the human response $Y_{\text{H}}$. 

Finally, we consider the setting where the query $Q$, the accurate context humans perceive $C_{\text{H}}$, and the human navigation strategy $N_{\text{H}}$ are specified as model input, which we refer to as the \textit{context formation and navigation} case. In this case, the LLM will generate the outcome
\begin{equation}\label{eq:navigation-simulator}
Y_{\text{CFN}} \sim f_{\text{LLM}}(Q, C_{\text{H}}, N_{\text{H}}).
\end{equation}
Because $f_{\textbf{LLM}}$ is close to $f_{\textbf{H}}$, the simulated outcome under \textit{context formation and navigation} , $Y_{\text{CFN}}$, should follow a similar distribution as the human outcome $Y_{\text{H}}$. For the rest of this paper, we will empirically demonstrate that for complex decision-making in multi-agent environments with strategic interdependence and endogenously generated beliefs, simulations based on \textit{context formation and navigation}  could robustly generate outcomes aligned with human behaviors.

\section{Simulation and Empirical Analysis}\label{sec:simulation_empirical}

To apply and empirically validate our proposed framework, we implement it across experimental settings drawn from complex decision-making environments characterized by strategic interdependence and endogenous belief formation among multiple agents. \citet{kremer2016inferring} provide a representative example of such an environment. We choose it as our focal case for two reasons. First, its main findings have been validated through human replications in the \textit{Management Science} Replication Project \citep{davis2023areplication}, which offers open-access data, instructions, and protocols that ensure transparency and comparability. Second, the study has recently been used in other LLM-simulated replication attempts (e.g., \citealp{kirshner2024artificial}).


\subsection{Overview of the Experimental Setting}\label{sec:exp_setting}

To provide the context for our framework application and evaluation, we briefly review the experimental setting of \citet{kremer2016inferring}. The paper examines how uninformed consumers infer product quality from queue length in a sequential purchasing environment. In this setting, uninformed consumers rely on observed wait times as product quality signals because queue length reflects the purchasing behavior of informed consumers. The study reports a main finding that the presence of informed consumers affects the impact of queue length on purchase decisions of uninformed consumers.

In the experiment, groups of four participants enter a market where a product is either of high value (\$3.50) or of low value (\$0), each with 50\% probability. Participants are randomly assigned either to be informed of the true product value or to remain uninformed with 50\% probability. Purchase decisions are made sequentially and, before making a decision, each participant observes the delivery wait time. If a participant decides to purchase, their delivery wait time equals one production period plus the number of participants who have purchased before them. Each participant's payoff equals the \$4-endowment plus the product value (if purchased), minus a waiting cost of \$1 for each unit of wait time.

While \citet{kremer2016inferring} originally study additional treatment conditions, following \citet{davis2023areplication}, we focus on the comparison between Q00 (no informed participants) and Q50 (50\% of participants are informed) conditions. In condition Q00, all participants are uninformed, so wait times carry no information about the product quality. In condition Q50, half of the participants are informed, so that their purchasing decisions make wait time an informative signal for uninformed participants to infer quality.

Following \citet{davis2023areplication}, we examine \citet{kremer2016inferring}'s finding that uninformed consumers interpret queue length as a quality signal using two hypothesis tests based on probit regressions. The first hypothesis (H1) examines whether uninformed consumers are less likely to purchase when they face an empty queue $(w = 1)$ in the presence of informed consumers. Here, $w$ denotes the delivery wait time, defined as the number of periods until the product is delivered; thus $w=1$ corresponds to an empty queue with no prior purchases. We estimate the following probit model:
\begin{equation}\label{eq:debo_h1}
   \mathbb{P}(\text{Purchase} = 1)
    = \Phi\!\left(\beta_0 + \sum_{w \in \{2,3,4\}} \beta_w D_w
    + \sum_{w \in \{1,2,3,4\}} \gamma_w D_{w} \times D_{Q50}\right)
\end{equation}
where, $\Phi(\cdot)$ is the CDF of the standard normal distribution, $D_w$ is a dummy variable equal to 1 if the delivery wait time is $w \in \{1,2,3,4\}$ and 0 otherwise, with $w=1$ omitted as the baseline category, $D_{Q50}$ is a dummy variable for condition Q50, and $D_w \times D_{Q50}$ captures the interaction between each wait-time dummy $w \in \{1,2,3,4\}$ and condition Q50. In this specification, the coefficient $\gamma_1$ on $D_1 \times D_{Q50}$ captures the change in purchase probability at $w=1$ when the condition changes from Q00 to Q50. Accordingly, H1 is empirically supported when $\gamma_1$ is negative and statistically significant.

The second hypothesis (H2) examines whether the negative effect of wait time on purchase is reduced when informed consumers are present. We estimate the interaction effect between wait time $w$ and treatment condition Q50 in the following probit model:
\begin{equation}\label{eq:debo_h2}
\mathbb{P}(\text{Purchase} = 1) = \Phi\left(\beta_0 + \beta_1 w + \beta_2 D_{Q50} + \beta_3  D_{Q50}\times w\right).
\end{equation}
where $w$ is the delivery wait time, $D_{Q50}$ is a dummy variable for condition Q50, and $D_{Q50} \times w$ captures the interaction between wait time and condition Q50. Accordingly, H2 is empirically supported when $\beta_3$ is positive and statistically significant.

\subsection{Simulation Design}\label{sec:simulation_design}

To replicate the experimental setting of \citet{kremer2016inferring} using LLM, we simulate cohorts of agents powered by OpenAI's GPT-4o snapshot from August 6th, 2024. The temperature parameter controlling stochasticity in the LLM output is set to its default value 1, allowing for variation across simulated samples. The agents do not share memory between each other. We adopt LangChain,\footnote{\url{https://www.langchain.com/}} an open-source framework for building agent-based systems, to instantiate and manage the LLM-simulated subjects. 


\citet{kremer2016inferring}'s original sample consists of 100 students assigned to 25 cohorts of 4 (8 cohorts in condition Q00 and 17 cohorts in condition Q50), each cohort completing 26 sequential decision-making rounds. \citet{kremer2016inferring} demonstrate that learning effects and round dependencies are minimal in their original study, so we implement the simulation as a between-subjects design with independently generated one-shot decisions. This allows us to take advantage of the LLM's ability to generate independent responses at scale while avoiding context-window overflow. We simulate 208 ($8 \times 26$) cohorts in condition Q00 and 442 ($17 \times 26$) cohorts in condition Q50, reproducing the same total number of experimental units as \citet{kremer2016inferring}. 


Within each cohort, LLM-simulated subjects are randomly assigned to positions in the decision sequence and to informed or uninformed roles. Once a subject makes a decision, the purchase outcome is recorded and determines the wait time presented to the next subject in that cohort, preserving the endogenous evolution of the queue. Across cohorts, simulations are conducted independently. Each run uses new role assignments and initializations for all subjects. Our approach preserves both the total number of observations and the causal structure of the original experiment, including the sequential nature of decisions and the endogenous formation of the wait line.

\subsection{Prompt Design} \label{sec:prompt_design}

\begin{figure}[ht]
\centering
\includegraphics[width=\textwidth]{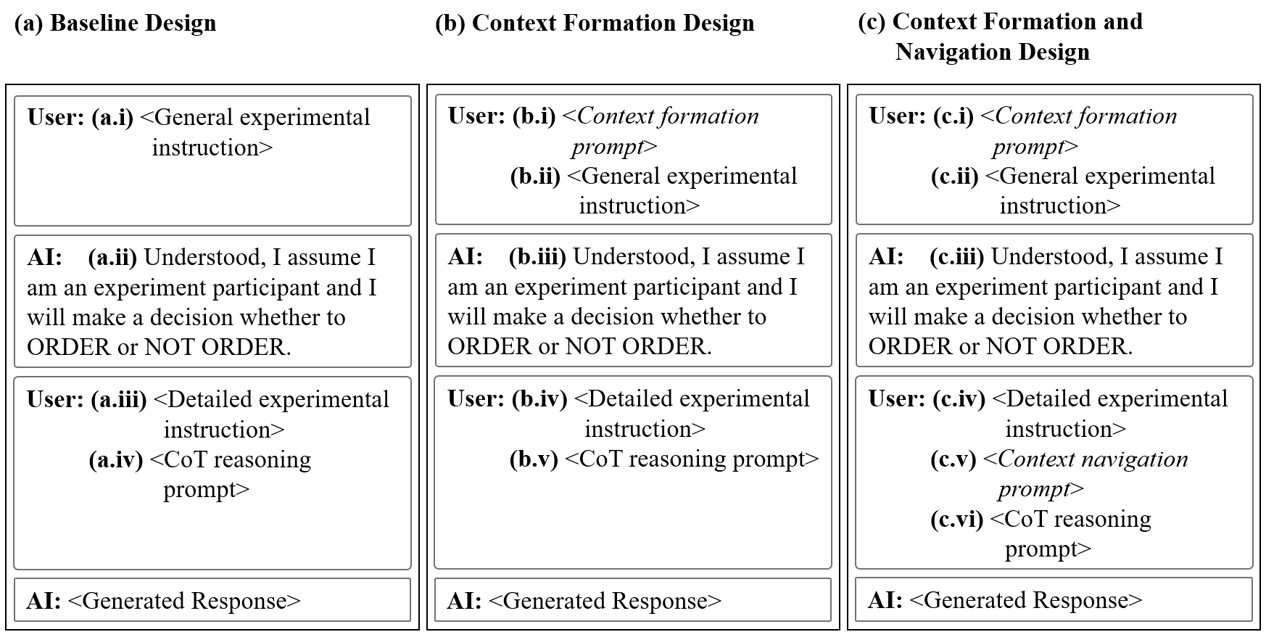}
\caption{Prompting Methods for Baseline, Context Formation, and Context Formation and Navigation}
\label{fig:one}
\end{figure}

Guided by our theoretical framework, we construct three prompt designs that vary in whether they include (i) neither \textit{context formation} nor \textit{context navigation}, which will generate the outcome $Y_{\text{Base}}$ by Eqn. \eqref{eq:base-simulator}, (ii) \textit{context formation} only, which will generate the outcome $Y_{\text{CF}}$ by Eqn \eqref{eq:formation-simulator}, or (iii) both components, which will generate the outcome $Y_{\text{CFN}}$ by Eqn. \eqref{eq:navigation-simulator}. Figure \ref{fig:one} panels (a), (b), and (c) present the prompt design of the \textit{baseline}, \textit{context formation}, and \textit{context formation and navigation} designs, respectively. These variants operationalize the two-stage structure of our framework and allow us to see how each stage contributes to alignment with human behavior. 

Figure \ref{fig:one}(a) reveals the \textit{baseline} design. Each LLM-simulated subject first receives a general experiment instruction adapted from the experiment instruction of \citet{kremer2016inferring} as part of the initialization, i.e., Figure \ref{fig:one}(a.i). Then, we append a brief compliance confirmation message to ensure the agent's behavior is under control \citep[Figure \ref{fig:one}(a.ii), see also][]{goli2024frontiers}. In each decision round, we append the detailed experimental instruction (e.g., wait time and, when applicable, product value) and a standard CoT reasoning prompt \citep[Prompt \ref{prompt:debo_cot_prompt}; see also][]{wei2022chain,kojima2022large}, as shown in Figure \ref{fig:one} (a.iii) and (a.iv), respectively. This prompting setup reflects a common approach in attempts to simulate human behavior with LLMs \citep{goli2024frontiers,kirshner2024artificial}.

\begin{promptbox}[title=Prompt \thepromptbox: CoT reasoning prompt for \citet{kremer2016inferring}]\label{prompt:debo_cot_prompt}

Put your step-by-step reasoning in \textless reason\textgreater\textless/reason\textgreater{} and the final decision (ORDER/NOT ORDER) in \textless answer\textgreater\textless/answer\textgreater.

\end{promptbox}

Guided by our framework, the \textit{context formation} design builds on the \textit{baseline} design by adding the \textit{context formation} prompt (Prompt \ref{prompt:debo_formation_prompt}) at the beginning as shown in Figure \ref{fig:one} (b.i), which explicitly describes the two experimental conditions to improve how LLMs understand the context of the task ($C_{\text{H}}$ in Eqn. \eqref{eq:formation-simulator}). While human participants are typically not informed of all experimental conditions in order to preserve ecological validity, leaving the experimental design implicit can be problematic in LLM social simulations. In particular, when the experimental design is not explicitly specified, LLM-simulated subjects may misinterpret the task context, leading to behavioral misalignment \citep{gui2025challenge}. Accordingly, to make the context more transparent, we incorporate the \textit{context formation} prompt below into the full prompts presented to all LLM-simulated subjects, regardless of the conditions they are assigned to.

\begin{promptbox}[title=Prompt \thepromptbox: Context formation prompt generated for \citet{kremer2016inferring}]\label{prompt:debo_formation_prompt}

This study consists of two different conditions, and each participant will be assigned to one of them at random. You will not switch between conditions during the study; you will only participate in one of the two conditions. The two conditions differ in the information available to consumers before they make their purchasing decisions.

\textbullet\ Condition 1 (Q00): In this condition, all consumers are uninformed about the true value of the product before making a purchase decision. Each product can either be a high-value product (\$3.50) or a low-value product (\$0) with equal probability (50\%). However, consumers will not know whether the product is high-value or low-value at the time of ordering.

\textbullet\ Condition 2 (Q50): In this condition, some consumers are informed while others remain uninformed. If you are assigned to this condition, you will first learn whether you are an informed or uninformed consumer. Informed consumers will know the product's true value before making a purchase decision, while uninformed consumers will not.

You are assigned to condition Q50/Q00. The following sections will provide the detailed instructions.

\end{promptbox}

The \textit{context formation and navigation} design includes both components, as shown by Figure \ref{fig:one}(c). It supports \textit{context formation} by clarifying the experimental setup at the beginning (Figure \ref{fig:one}(c.i)) and \textit{context navigation} by adding an instruction in each round that directs the model to selectively reason within the given context ($N_{\text{H}}$ in Eqn. \eqref{eq:navigation-simulator}, Figure \ref{fig:one}(c.v), and Prompt \ref{prompt:debo_navigation_prompt}).

\begin{promptbox}[title=Prompt \thepromptbox: Context navigation prompt generated for \citet{kremer2016inferring}]\label{prompt:debo_navigation_prompt}

Reason in the following approach: Hold an initial belief about what is unobserved, consider what the observable cue suggests, update your belief with a mix of reasoning and intuition, and make your decision accordingly.

\end{promptbox}


\subsection{Simulation Results} \label{sec:simulation_results}

Following \citet{davis2023areplication}, we evaluate replication along two hypotheses: H1 assesses whether uninformed consumers are less likely to purchase in condition Q50 than in condition Q00 when $w = 1$, and H2 tests whether the negative effect of wait time on purchasing is attenuated in condition Q50 relative to condition Q00. Based on our theoretical framework, we expect that the \textit{baseline} design is unlikely to reproduce these human behavioral patterns, the \textit{context formation} design can improve alignment, and the \textit{context formation and navigation} design is expected to fully replicate the original results. 



We present the simulation results in Table \ref{tab:one}. Under our \textit{baseline} design, which uses the original human instructions together with a standard CoT reasoning prompt, we fail to replicate either H1 or H2. For H1, the estimated $\gamma_1$ is marginally significant but in the wrong direction ($p = .059$, $\hat\gamma_1 = 0.59$). For H2, the interaction term between wait time and Q50 condition, $\beta_3$, is statistically significant but negative ($p = .05$, $\hat\beta_3 = -0.97$). The CoT reasoning traces show that the uninformed LLM-simulated subjects in both conditions follow a simple payoff-comparison rule without accounting for the quality-signaling effect of queue length. They calculate the expected value of ordering as EV = 4 $-$ wait time $+$ 1.75, where 1.75 is the expected product value under the 50/50 prior and 4 is the endowment, and compare this expected value directly with the guaranteed endowment from not ordering. Accordingly, subjects in both conditions make identical decisions across wait times. Purchasing is only preferred when $\text{EV} > 4$, which implies $w < 1.75$. As a result, subjects purchase only at the minimal wait time ($w = 1$) and not at longer wait times ($w > 1$), regardless of the condition they are in. The LLM-simulated subjects do not refer to others' actions or to informational cues of the wait time in reasoning, which explains the joint replication failure of H1 and H2.




The \textit{context formation} design replicates H1 in the statistical sense ($p =0.04$, $\hat\gamma_1 = -0.77$), implying the lower likelihood of purchase in Q50 than in Q00 for $w=1$. In terms of H2, LLMs stop purchasing entirely once $w>1$ in both conditions. This produces no variations in purchasing for $w >1$, making it impossible to estimate how the effect of wait time differs across conditions. Although this prevents a formal statistical test of H2, the pattern already diverges from human behavior, where the purchase rate decreases with wait time but remains strictly above zero in both conditions.



A closer look at the CoT reasoning traces under \textit{context formation} reveals additional insights. Uninformed LLM-simulated subjects, albeit with improved awareness of their own uninformed states, fail to recognize that some consumers are informed, so that no one waiting (i.e., $w=1$) signals the low quality of the product. Uninformed consumers explicitly note their own status and become more risk-averse if they know there are informed ones on the market, leading to a lower purchase likelihood under Q50. Furthermore, their decision rules remain the same as the \textit{baseline} case, i.e., the subjects directly compute the expected value of ordering and compare it with the endowment, without drawing any inference from the observed wait time. Because they do not interpret the queue length as conveying information about quality, their decisions remain constant (i.e., not purchase) for all $w>1$. This lack of variation not only prevents the estimation of $\beta_3$ for H2, but also reflects behavioral misalignment, as humans continue to make purchases at longer waits. Thus, the \textit{context formation} design improves context awareness of the task but does not induce the model to infer quality from wait time, leaving the hypotheses unreplicated.




The \textit{context formation and navigation} design successfully replicates both H1 and H2 (H1:$p < 0.01$, $\hat\gamma_1 = -0.42$; H2: $p < 0.01$, $\hat\beta_3 = 1.60$), showing that combining \textit{context formation} and \textit{context navigation} achieves alignment. Under the \textit{context formation and navigation} design, the uninformed LLM-simulated subjects not only recognize the context but also reason about product quality based on wait time. The CoT reasoning traces show that the LLM-simulated subjects refer to earlier consumers' actions as cues to adjust their beliefs about product quality before deciding whether to purchase. As a consequence, the model moves from a fixed payoff-comparison rule purely based on expected-value calculation to a context-sensitive reasoning process in a dynamic environment.


This improved navigation of the context yields LLM behaviors more closely aligned with human decisions. The LLM-simulated subjects order less often at short waits when no one purchases, replicating the lower initial purchasing rate in condition Q50 (H1). They also maintain moderate purchasing at longer waits when earlier purchases suggest high product quality, reproducing the attenuated wait-time effect observed in human experiments (H2). Thus, the \textit{context formation and navigation} design not only improves the model's context awareness, but also guides its reasoning within that context,  successfully replicating both hypotheses in \citet{kremer2016inferring}. 


A potential concern of our results is that successful replication simply reflects memorization, i.e., the original study may appear in the GPT-4o's training data. If memorization alone were driving replication, the baseline design, using the same underlying model but without additional prompt design, should also reproduce the original results of \cite{kremer2016inferring}. The failure of the \textit{baseline} design, therefore, indicates that replication success under the \textit{context formation and navigation} design should be attributed to our prompt design rather than memorization by LLM.

Our framework can be applied to different settings. For replicating existing studies, \textit{context formation} prompts explicitly specify the experimental design based on the original instructions (e.g., treatment conditions, payoff structures, decision variables), ensuring all elements are represented accurately. \textit{Context navigation} prompts can be derived from the study's core theoretical mechanism (e.g., the authors' explanation for how their treatment works). For new experimental studies not reported in the literature, \textit{context formation} follows the same generation process, but \textit{context navigation} requires an iterative approach. Because new experiments are designed to test whether a proposed economic or behavioral mechanism is valid, navigation prompts cannot directly specify how participants reason. Instead, they should be grounded in evidence from pilot studies (e.g., think-aloud protocols for humans) or the theoretical mechanism under investigation, and refined iteratively as empirical patterns emerge. Section~\ref{sec:applications} demonstrates the framework's application in two additional experimental settings.


\begin{table}[ht]
\centering\scriptsize
\caption{Human and LLM Replications of \citet{kremer2016inferring}}
\label{tab:one}

\resizebox{\textwidth}{!}{
\begin{tabular}{
  l
  l
  >{\centering\arraybackslash}m{3.2em}
  r
  r
  c
  >{\centering\arraybackslash}m{5em}
  r
  r
  c
}
\\[-1.8ex]\hline
\hline\\[-1.8ex]
\multicolumn{2}{c}{} &
  \multicolumn{4}{c}{\textbf{Hypothesis 1}} &
  \multicolumn{4}{c}{\textbf{Hypothesis 2}} \\
\cmidrule(lr){3-6} \cmidrule(lr){7-10}
\textbf{Paper/Prompt Design} & \textbf{Subject Type} &
$\gamma_1$ & $\chi^2$ & $p$-value & \textbf{Replication Success} &
$\beta_3$ & $\chi^2$ & $p$-value & \textbf{Replication Success} \\
\hline\\[-1.8ex]
\citet{kremer2016inferring} & Human &
\makecell{-1.14} & 31.23 & $<0.01$ & \makecell{--} &
\makecell{0.85} & 74.34 & $<0.01$ & \makecell{--} \\
\citet{davis2023areplication} (South Carolina) & Human &
\makecell{-0.84} & 22.72 & $<0.01$ & \makecell{\cmark} &
\makecell{0.22} & 5.35 & 0.021 & \makecell{\cmark} \\
\citet{davis2023areplication} (Michigan) & Human &
\makecell{-1.09} & 47.94 & $<0.01$ & \makecell{\cmark} &
\makecell{0.69} & 57.72 & $<0.01$ & \makecell{\cmark} \\
Baseline & LLM &
\makecell{0.59} & 3.79 & 0.05 & \makecell{\xmark} &
\makecell{-0.97} & 4.02 & 0.05 & \makecell{\xmark} \\
Context Formation & LLM &
\makecell{-0.77} & 4.43 & 0.04 & \makecell{\cmark} &
\makecell{--} & \makecell{--} & \makecell{--} & \makecell{\xmark} \\
Context Formation and Navigation & LLM &
\makecell{-0.42} & 7.22 & $<0.01$ & \makecell{\cmark} &
\makecell{1.60} & 43.53 & $<0.01$ & \makecell{\cmark} \\
\hline
\hline\\[-1.8ex]
\end{tabular}%
}

\begin{tablenotes}
\item \textit{Note}: For simulation results of the \textit{context formation} design, probit regression for Hypothesis 2 can not be performed due to uniform behavior: LLMs ceased purchasing entirely at any wait time > 1 in both conditions, making wait time a perfect predictor of purchase decision and thus producing no variation for estimating the interaction effect. This rigid behavior contrasts with the gradual decline in purchasing observed in human data, suggesting that \textit{context formation} alone fails to simulate human-like variability in decision-making.
\end{tablenotes}
\end{table}

\section{Applications of the Theoretical Framework}
\label{sec:applications}

In this section, we apply the two-stage framework beyond \citet{kremer2016inferring} to evaluate its generalizability across distinct decision-making environments and across LLM models. Section \ref{sec:gui_toubia} examines a documented LLM alignment failure in consumer demand estimation reported by \citet{gui2025challenge}, serving as a benchmark for assessing the framework in a simple decision task that lacks the strategic interdependence and endogenous belief formation of our focal applications. In Section \ref{sec:cason}, we extend our framework to a crowdfunding game characterized by strategic interdependence and endogenously generated beliefs from \citet{cason2024signaling}. Section \ref{sec:cross_model} tests whether the patterns observed in our three experiments generalize across several SOTA LLMs by applying a common prompt design to each model and comparing their alignment outcomes.


\subsection[Application to Gui and Toubia (2025): Demand Estimation]{Application to \citet{gui2025challenge}: Demand Estimation with Price Experiments}\label{sec:gui_toubia}


We begin by applying our theoretical framework to the demand-estimation experiment of \citet{gui2025challenge}, a simple  decision-making environment in which LLM-simulated subjects nonetheless fail to reproduce the downward-sloping human demand curve. In this study, 1{,}000 Prolific respondents each evaluate 40 consumer packaged goods whose prices are randomly drawn from 0\% to 200\% of the product's regular price. For each product-price pair, participants answer a binary question (``Would you or would you not purchase this product?''), and their responses are aggregated to estimate purchase probabilities at each price level. Consistent with the law of demand, human purchase probabilities decrease monotonically with price.


In \citet{gui2025challenge}'s LLM benchmark case, they present GPT-4o-mini with the same product-price combinations and generate 50 independent purchase decisions for each combination. These simulated decisions are aggregated in the same way as the human responses to produce purchase probabilities at each price level. The resulting demand curve deviates sharply from the human benchmark. Rather than the purchase probabilities decreasing in price, the model produces an inverted-U shape curve in which purchase likelihood first increases and then decreases with price \citep[Figure 2 in][]{gui2025challenge}. Such misalignment motivates applying our framework to this simple, non-strategic setting. 
\subsubsection{Prompt Design}\label{Gui_2025_prompt}

The main body of the \textit{baseline} design follows the prompt design of \citet{gui2025challenge}, with a CoT reasoning prompt appended at the end. The \textit{context formation} design is built on the \textit{baseline} design by adding an LLM-generated introduction that clarifies the experimental setting at the beginning, following the same prompt-generation procedure introduced in Section~\ref{sec:prompt_design}.

\begin{promptbox}[title=Prompt \thepromptbox: Context formation prompt generated for \citet{gui2025challenge}]\label{prompt:gui_formation_prompt}

In this study, each participant is presented with one of 40 different products, shown at a price drawn from 11 systematically varied levels ranging from 0\% to 200\% of its regular retail value. For each product-price scenario, participants provide a single binary response - whether they would purchase the product or not. 
You are assigned a survey question in which the price level of the product is manipulated.

\end{promptbox}

The \textit{context navigation} prompt guides the model to selectively reason within the given context by focusing on features that are relevant for decision-making.

\begin{promptbox}[title=Prompt \thepromptbox: Context navigation prompt generated for \citet{gui2025challenge}]\label{prompt:gui_navigation_prompt}

Reason in the following approach: Notice what stands out in the context, use it to guide your choice, and don't overcomplicate with unnecessary details.

\end{promptbox}


\subsubsection{Simulation Results}

\begin{figure}[!t]
    \centering
    \includegraphics[width=\linewidth]{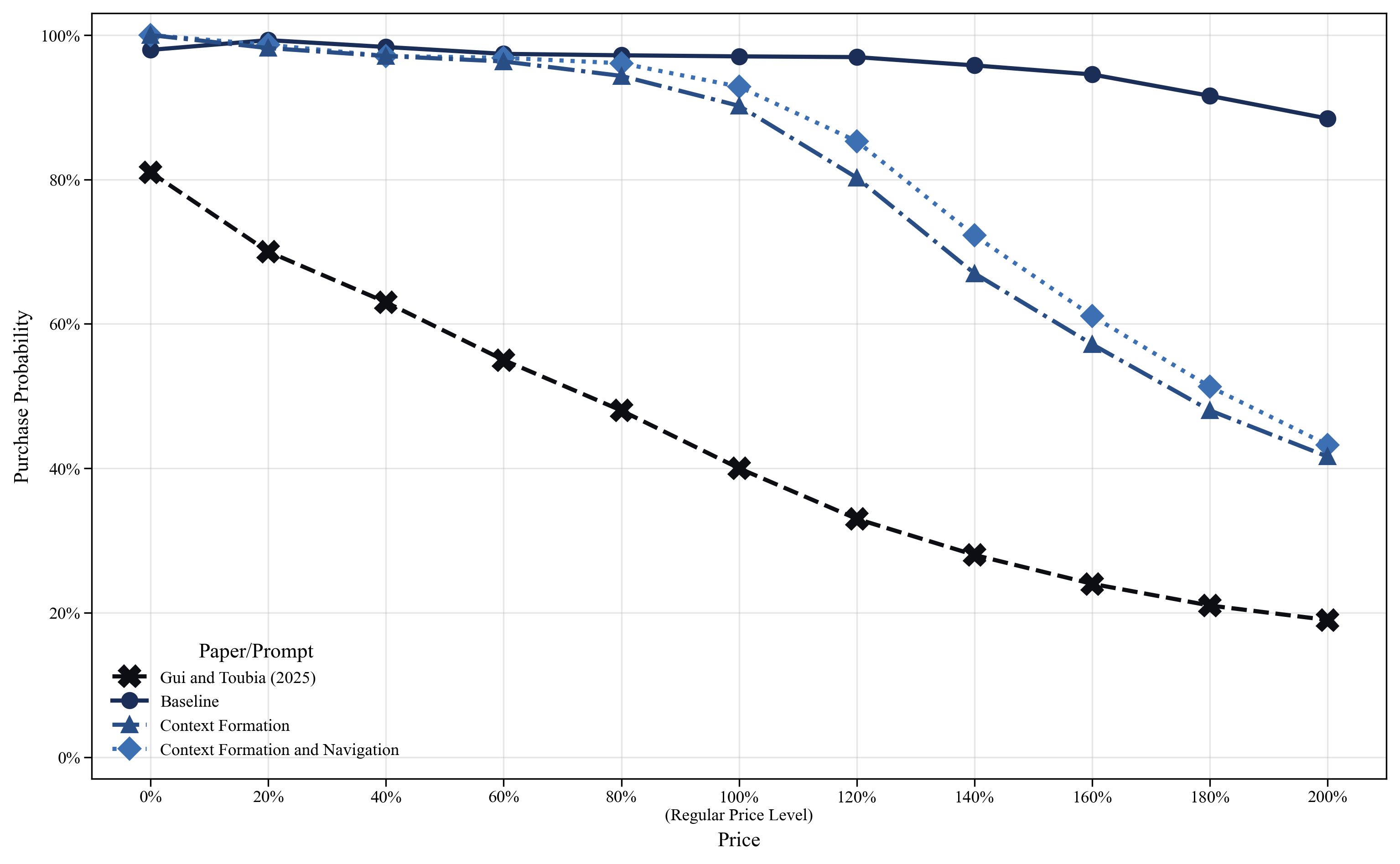}
    \caption{Demand curve elicited from LLMs (GPT-4o)}
    \label{fig:demand_curve}
        \begin{tablenotes}
            \fontsize{9}{11}\selectfont
            \itshape
            \item Note: The ``Gui and Toubia (2025)'' line is an approximation of the human demand curve reported in Gui and Toubia (2025, Fig.~2) and is not reproduced from the original human survey data; small discrepancies may arise due to figure resolution.
          \end{tablenotes}  
\end{figure}

To evaluate simulation performance, \autoref{fig:demand_curve} plots average purchase probabilities across relative price levels under each prompt design and compares the resulting demand curves to the human benchmark reported by \citet{gui2025challenge}. Using the more advanced GPT-4o \citep[rather than GPT-4o-mini as in][]{gui2025challenge} and appending an additional CoT prompt to our \textit{baseline} design, the \textit{baseline} simulations reproduce the inverted-U pattern reported in \citet{gui2025challenge}'s LLM-based simulations. However, this simulation pattern diverges from the human benchmark, which exhibits a clearly downward-sloping relationship. Examination of the CoT reasoning traces suggests that, under the \textit{baseline} design, LLM-simulated subjects interpret the zero-price level (0\% of the regular price) as a potential input or pricing error and therefore choose not to purchase. This behavior indicates that the model focuses on the anomalous nature of a zero price rather than reasoning about price in a standard consumer demand context.

Meanwhile, the \textit{context formation} design produces a steadily declining demand curve, with purchase probabilities starting close to 100\% at the lowest prices and falling to around 40\% at the highest. This reproduces a fundamental principle of consumer behavior: the law of demand. Examination of the CoT reasoning traces shows that the \textit{context formation} design changes how the LLM-simulated subjects interpret prices. Under this design, a price of zero is interpreted as the product being genuinely free rather than as a suspicious abnormality with hidden conditions, which in turn yields a downward-sloping demand curve aligned with human behavior in the original study. By contrast, while the \textit{context formation and navigation} design also manages to generate a downward-sloping demand curve, it does not yield clear improvements over the \textit{context formation} design alone. 


These results indicate that, in the demand estimation setting, \textit{context formation} is sufficient to replicate human behavior. Because the task is simple, featuring binary purchase decisions with price as the only manipulated variable and no strategic interaction or belief formation, the model does not require explicit guidance on how to reason through the decision process. Explicitly framing the randomized design through \textit{context formation} is therefore sufficient for the model to arrive at the appropriate reasoning process on its own. This contrasts with the \citet{kremer2016inferring} replication, where the strategic interdependence and endogenous belief formation required both \textit{context formation} and \textit{context navigation} to achieve behavioral alignment. It is worth noting that \citet{gui2025challenge} also explores an alternative simulation design to align with human decision patterns. Both \citet{gui2025challenge}'s design and our \textit{context formation} design aim to enhance LLMs' understanding of the problem by making price manipulation and variation explicit. In essence, unblinding is an implementation of \textit{context formation} in our framework.


\subsection[Application to Cason et al. (2025): Refund Bonus and Information Asymmetry in Crowdfunding]{Application to \citet{cason2024signaling}: Refund Bonus and Information Asymmetry in Crowdfunding}\label{sec:cason}

In this section, we extend our framework to the crowdfunding experiment of \citet{cason2024signaling}, which examines how refund bonuses mitigate information asymmetries in a strategically interdependent investment environment. The setting features two inventors and two investors interacting in a repeated game. In each round, each inventor offers a single invention of fixed quality, either good or bad, while each investor, endowed with \$2, decides independently whether to invest \$1 in each of the two inventions.

Inventors observe their own invention's quality, whereas investors do not. Invention quality is good with prior probability 0.4286. For bad inventions, investors receive a negative report with 25\% probability that fully reveals low quality. If no negative report is received, the invention is equally likely to be good or bad. An invention is funded only if both investors invest. If funded, good inventions yield \$3 per investor, while bad inventions result in a loss of the \$1 investment. Uninvested funds can be placed in a safe outside option yielding \$1.50.

The experiment studies whether refund bonuses, payments promised to investors if an invention fails to reach the funding threshold, can serve as a signal of quality. Specifically, if only one investor invests in an invention, that investor receives a \$1 refund bonus in addition to the returned \$1 investment. Because bad inventions are more likely to trigger bonus payments, offering a refund bonus is more costly for low-quality inventions, making bonuses a potential separating signal. Experimental conditions differ in how bonuses are determined. In the Endogenous Bonus (EB) condition, inventors choose whether to offer a bonus after observing quality. In the Random Bonus (RB) condition, bonuses are assigned randomly and independently of quality. The main finding is that good inventions are successfully funded more often in the EB condition than in the RB condition.

\citet{cason2024signaling}'s original design assigns 168 subjects to 14 matching groups of 12, with 8 groups in the EB condition and 6 groups in the RB condition, where each subject repeatedly performs decision-making for 40 rounds. Taking advantage of the LLM's ability to simulate subjects at scale, we modify the design by increasing the number of subjects and reducing the number of decision rounds per subject. We simulate 40 independent groups of 12 subjects in each condition, with each subject completing 8 decision rounds. 

To evaluate whether the main finding of \citet{cason2024signaling} can be replicated, we compare funding success for good-quality projects that offer refund bonus in the EB and RB conditions using a Wilcoxon rank-sum test. We treat whether funding success is systematically higher the EB condition than in the RB condition. A positive and statistically significant z-statistic indicates successful replication of the main finding.

\subsubsection{Prompt Design} \label{sec:cason_prompt}

We use the same approach to construct prompt design as in the previous \citet{kremer2016inferring} replication. The \textit{baseline} prompt consists of the original experiment instruction and a CoT reasoning prompt added at the very end. Consistent with the approach introduced in Section~\ref{sec:prompt_design}, the \textit{context formation} design is again implemented by adding an LLM-generated overview of the experimental conditions.

\begin{promptbox}[title=Prompt \thepromptbox: Context formation prompt generated for \citet{cason2024signaling}]\label{prompt:cason_formation_prompt}

In this experiment, you will participate in a series of decision-making rounds where people take on the roles of investors and inventors. Each round involves two inventions that may receive funding if both investors choose to support them. Investors decide where to invest based on limited information, and inventors may have the option to offer a refund bonus---a payment to compensate investors if the invention fails to get enough support.

There are two experimental conditions that differ in how refund bonuses are determined. In the Endogenous Bonus (EB) condition, inventors choose whether to offer a refund bonus after learning the quality of their invention. In the Random Bonus (RB) condition, refund bonuses are assigned randomly with a 50\% chance, regardless of invention quality. You have been randomly assigned to one of these two conditions.

You are assigned to the Random/Endogenous Bonus condition. The following sections will provide the detailed instructions.

\end{promptbox}


Because \citet{cason2024signaling} studies a strategic costly-signaling game with asymmetric information, the \textit{context navigation} prompt functions as an instruction that structures how the model reasons within the given context.

\begin{promptbox}[title=Prompt \thepromptbox: Context navigation prompt generated for \citet{cason2024signaling}]\label{prompt:cason_navigation_prompt}

Use the following approach: explicitly state priors, likelihoods, compute the posterior via Bayes' rule, then compare expected values of both actions before deciding.

\end{promptbox}

\subsubsection{Simulation Results} \label{sec:results_cason}

\autoref{tab:two} reports the replication performance of the main result in \citet{cason2024signaling} across the three prompt designs. Consistent with our earlier replication of \citet{kremer2016inferring}, neither the \textit{baseline} nor the \textit{context formation} designs could replicate the main result. In contrast, the \textit{context formation and navigation} design successfully replicate the behaviors observed in human experiments. Furthermore, we investigate the qualitative patterns in the CoT reasoning traces for each prompt design, offering insight into how different prompt designs affect simulation results in this complex decision-making environment.

Under the \textit{baseline} design, we do not replicate the main finding. \autoref{tab:two} shows a statistically significant effect in the opposite direction ($z = -4.27$, $p < 0.01$): good inventions are funded less often in the EB condition than in the RB condition. The CoT reasoning traces suggest that this reversal is driven by inventors' overreliance on the negative report. Many inventors infer, incorrectly, that good inventions never generate negative reports and therefore will attract more investors without needing a refund bonus. As a result, they treat the report, rather than the refund bonus, as the primary determinant of funding and largely ignore the bonus as an incentive mechanism. In parallel, investors in both EB and RB conditions apply a similarly shallow heuristic: they compare the payoff values in the instructions and interpret the refund bonus as simply additional money, rather than using it to infer underlying invention quality. Consistent with these patterns, inventors often view the refund bonus as a potential cost and thus rarely offer it, yielding a low bonus-offer rate for good inventions in the EB condition, whereas in the RB condition good inventions receive bonuses at the exogenously assigned rate of 50\%. Because investors respond positively to bonuses regardless of their source, good inventions receive more bonus offers, and therefore higher funding rates, in RB than in EB, producing the reversed result in the simulation.

Likewise, the \textit{context formation} design still yields a statistically significant effect in the wrong direction ($z = -5.06$, $p < 0.01$). Although this design clarifies how refund bonuses are determined across conditions, it does not meaningfully change the underlying reasoning. Inventors remain anchored to the negative report signal and continue to neglect the refund bonus as an incentive mechanism. Investors likewise show no deeper inference, treating bonuses simply as higher payoffs. Consequently, both the reasoning patterns and behavioral outcomes under this design are largely indistinguishable from those observed under the \textit{baseline} case.

With the \textit{context formation and navigation} design, we successfully replicate the main finding of \citet{cason2024signaling} ($z = 2.45$, $p = 0.01$). The \textit{context navigation} prompt directs attention to the key relationships among invention quality, the refund bonus decision, and investor response. Inventors become less anchored to the negative report signal and begin to view the refund bonus as a mechanism for influencing investor behavior. In parallel, both inventors and investors shift from comparing static payoff entries to evaluating expected payoffs. This change mitigates the earlier distraction induced by the negative report and yields investment patterns consistent with the human benchmark, reproducing the result that good inventions are funded more often in the EB condition than in the RB condition.

Taken together, our results indicate that both \textit{context formation} and \textit{context navigation} are necessary to replicate \citet{cason2024signaling}. In contrast to the simpler demand-estimation setting, where \textit{context formation} design alone suffices, the strategic interdependence and belief-driven nature of crowdfunding requires both components to align the LLM-simulated subjects' behavior with human outcomes.

\begin{table}[ht]
\centering\scriptsize
\caption{LLM Replication Results for \citet{cason2024signaling}}
\label{tab:two}

\resizebox{0.7\textwidth}{!}{%
\begin{tabular}{ll r r c}
\\[-1.8ex]\hline
\hline\\[-1.8ex]
\textbf{Paper/Prompt Design} & \textbf{Subject Type} &
$z$ & $p$-value & \textbf{Replication Success} \\
\hline\\[-1.8ex]
\citet{cason2024signaling} & Human &
3.10 & $<0.01$ & \makecell{--} \\
Baseline & LLM &
-4.27 & $<0.01$ & \makecell{\xmark} \\
Context Formation & LLM &
-5.06 & $<0.01$ & \makecell{\xmark} \\
Context Formation and Navigation & LLM &
2.45 & 0.01 & \makecell{\cmark} \\
\hline
\hline\\[-1.8ex]
\end{tabular}%
}

\begin{tablenotes}
\item \textit{Note}: $z$ statistics and asymptotic $p$-values are based on the tests of the corresponding results reported in \citet{cason2024signaling} and in the LLM simulations.
\end{tablenotes}
\end{table}

\subsection{The Impact of LLM Choice}\label{sec:cross_model}

In this section, we demonstrate the generalizability and robustness of our framework by implementing it with different LLMs to simulate the three experiments studied in this paper. In particular, we seek to uncover insights on how SOTA LLMs with reasoning capability may affect the validity and performance of our framework. Specifically, we test GPT-5 snapshot from Aug. 7th 2025, Claude 4.0-Sonnet-Thinking from May 14th, 2025, and DeepSeek-R1 from May 28th, 2025 under the same experimental settings, simulation procedures, and evaluation criteria while setting all their temperature parameters to the default value 1. We report the simulation results of all models and experiments in \autoref{tab:three}.

We highlight three key observations. First, across all models and all three studies, the \textit{context formation and navigation} design successfully replicates the original human results. This pattern suggests that providing both \textit{context formation} and \textit{context navigation} reliably aligns LLM behavior with human data, regardless of the model's baseline reasoning ability. Second, models with stronger reasoning abilities often succeed with \textit{context formation} alone. Such models (e.g., GPT-5) benefit from process-level supervision and verifier-style reinforcement learning that encourages more structured and internally coherent reasoning \citep[e.g.,][]{guo2025deepseek}, reducing their reliance on explicit navigation instructions and enabling them to effectively operate within the task representation supplied by \textit{context formation}. This pattern indicates that, when the context is made explicit, stronger models can often supply the appropriate navigation path endogenously. Finally, we observe a statistical replication of \citet{cason2024signaling} under the \textit{baseline} design with Claude-4.0-Sonnet-Thinking. However, its CoT reasoning traces suggest that the underlying reasoning still diverges from human decision processes in this setting, implying that \textit{context formation} remains necessary even for strong reasoning models.

Our cross-model results validate that our two-stage framework aligns LLM behaviors with human decisions when both \textit{context formation} and \textit{context navigation} are provided, while also showing that the need for \textit{context formation} depends on the model's reasoning ability. Nonetheless, even strong reasoning models sometimes require both stages to reproduce human patterns. Overall, the framework both improves behavioral alignment and clarifies when external context engineering is necessary versus when the model's internal reasoning suffices. More broadly, these implications connect to the industry notion of context engineering: shaping and structuring the information in an LLM's context window to guide task processing and reasoning. Our results suggest that as models' reasoning strength increases, alignment depends increasingly on the quality of the engineered contextual representation, which is implemented in our study through \textit{context formation}.

\begin{table}[ht]
\centering\scriptsize
\caption{Overview of Cross-Model and Cross-Design Replication Results}
\label{tab:three}

\renewcommand{\arraystretch}{1.2}
\resizebox{\textwidth}{!}{%
\begin{tabular}{l l c c c}
\\[-1.8ex]\hline
\hline\\[-1.8ex]
\textbf{Study} & \textbf{Model Name} &
\textbf{Baseline} & \textbf{Context Formation} & \textbf{Context Formation and Navigation} \\
\hline\\[-1.8ex]
\multirow{4}{*}{Kremer and Debo (2016)} &
GPT-4o                     & \xmark & \xmark & \cmark \\
& GPT-5                      & \xmark & \cmark & \cmark \\
& Claude-4.0-Sonnet-Thinking & \xmark & \xmark & \cmark \\
& DeepSeek-R1                & \xmark & \cmark & \cmark \\
\hline\\[-1.8ex]
\multirow{5}{*}{Gui \& Toubia (2025)} &
GPT-4o-mini                & \xmark & \cmark & \cmark \\
& GPT-4o                     & \xmark & \cmark & \cmark \\
& GPT-5                      & \xmark & \cmark & \cmark \\
& Claude-4.0-Sonnet-Thinking & \xmark & \cmark & \cmark \\
& DeepSeek-R1                & \xmark & \cmark & \cmark \\
\hline\\[-1.8ex]
\multirow{3}{*}{Cason et al. (2025)} &
GPT-4o                     & \xmark & \xmark & \cmark \\
& GPT-5                     & \xmark & \cmark & \cmark \\
& Claude-4.0-Sonnet-Thinking & \cmark & \cmark & \cmark \\
& DeepSeek-R1                & \xmark & \cmark & \cmark \\
\hline
\hline\\[-1.8ex]
\end{tabular}%
}
\end{table}

\section{Conclusion}\label{sec:conclusion}

This paper develops and tests a two-stage framework for improving behavioral alignment in LLM social simulations by separating \textit{context formation} (making the task representation explicit) from \textit{context navigation} (guiding how the model reasons within that representation). Across three established empirical settings and multiple frontier models, the results show a consistent pattern: providing both stages reliably recovers human behavioral findings in strategically interdependent, belief-driven environments, while stronger reasoning models can often succeed with \textit{context formation} alone when the task is simpler. The framework thus offers practical, theory-grounded ``context engineering'' guidance for designing, diagnosing, and evaluating LLM-simulated subjects as credible substitutes or complements to human subjects in complex decision-making environments.

The paper's main limitations are scope and mechanism. Empirically, it tests only three behavioral experiments, which may not span the broader space of dynamic, organizational, or information-rich settings where misalignment can manifest differently. Methodologically, it focuses on prompt-based interventions rather than training-time or architectural approaches (e.g., fine-tuning or RLHF-style methods), which may limit how broadly the framework transfers beyond prompting. Finally, the evaluation cannot fully keep pace with the most recent model releases. While our validation uses published studies that may appear in training data, our framework provides systematic guidance for future novel experiments where memorization is impossible.


Future work can extend this framework to richer socio-technical environments, e.g., multi-step decisions, group settings, and organizational workflows, where context evolves and interaction histories matter. It can also leverage the framework for systematic hypothesis generation, pretesting, and stress-testing theories before deploying costly human-subject studies. Finally, an important next step is to translate the framework into an actionable recipe, a set of concrete design choices, diagnostics, and reporting standards, that enables researchers and practitioners to apply \textit{context formation} and \textit{context navigation} systematically when building, auditing, and scaling LLM social simulations.
\footnote{An online appendix is available upon request.}

\bibliographystyle{informs2014}
\bibliography{reference}

\end{document}